\documentclass{article}
\usepackage{ijcai24}

\usepackage{times}
\usepackage{soul}
\usepackage{url}
\usepackage[hidelinks]{hyperref}
\usepackage[utf8]{inputenc}
\usepackage[small]{caption}
\usepackage{graphicx}
\usepackage{amsmath}
\usepackage{amsthm}
\usepackage{booktabs}
\usepackage{algorithm}
\usepackage{algorithmic}
\usepackage[switch]{lineno}
\usepackage{color}

\usepackage{multirow} 
\usepackage{amsfonts}
\usepackage{amsmath}  
\usepackage{graphicx}  
\usepackage{threeparttable}  
\usepackage[capitalize]{cleveref}
\crefname{section}{Sec.}{Secs.}
\Crefname{section}{Section}{Sections}
\Crefname{table}{Table}{Tables}
\crefname{table}{Tab.}{Tabs.}
\crefname{figure}{Fig.}{Figs.}

\title{CTAGE: Curvature-Based Topology-Aware Graph Embedding \\
for Learning Molecular Representations}

\author{
    Author Name
    \affiliations
    Affiliation
    \emails
    email@example.com
}

\begin{document}

\maketitle

\begin{abstract}
AI-driven drug design relies significantly on predicting molecular properties, which is a complex task. In current approaches, the most commonly used feature representations for training deep neural network models are based on SMILES and molecular graphs. While these methods are concise and efficient, they have limitations in capturing complex spatial information. Recently, researchers have recognized the importance of incorporating three-dimensional information of molecular structures into models. However, capturing spatial information requires the introduction of additional units in the generator, bringing additional design and computational costs. Therefore, it is necessary to develop a method for predicting molecular properties that effectively combines spatial structural information while maintaining the simplicity and efficiency of graph neural networks.

In this work, we propose an embedding approach CTAGE, utilizing $k$-hop discrete Ricci curvature to extract structural insights from molecular graph data. This effectively integrates spatial structural information while preserving the training complexity of the network. Experimental results indicate that introducing node curvature significantly improves the performance of current graph neural network frameworks, validating that the information from k-hop node curvature effectively reflects the relationship between molecular structure and function.
\end{abstract}

\section{Introduction}
Drug development is a lengthy, costly, and intricate process, involving drug discovery, clinical trials, and production approval. 
In recent years, deep learning-based molecular property prediction methods using data represented in SMILES~\cite{1984JOURNAL} strings have gained attention, for their potential to assist in drug discovery. 
Natural language processing (NLP) techniques have been applied to directly handle molecular SMILES, treating molecule generation as a Seq2Seq problem~\cite{2019N}. 
However, these methods disregard the natural topology of molecules and are insufficient for analyzing molecular data with temporal models alone. 

\begin{figure*}[htbp]
 \centering
 \includegraphics[width=13cm]{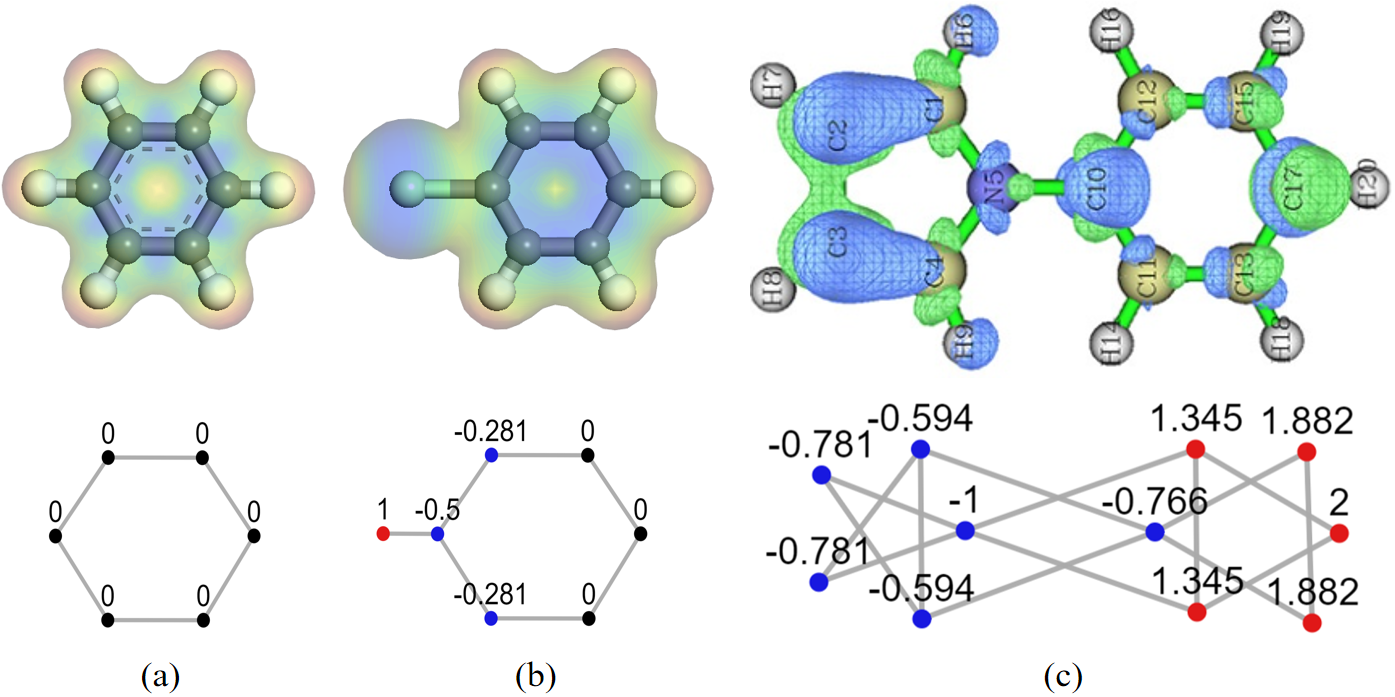}
 \caption{Electronic density maps: (a) benzene, (b) chlorobenzene, (c) differential electronic density map of 1-phenylpyrrole. The 1-hop node curvature information describes the density differences of electron clouds between benzene and chlorobenzene, while the 2-hop node curvature information of 1-phenylpyrrole describes more complex differences in differential electronic density.}
\label{Figure:00}
\end{figure*}

The graph convolutional networks improve the competitiveness of molecular modeling tasks by aggregating information from adjacent nodes and edges to update node features~\cite{2017Neural,kearnes2016molecular}. However, this mechanism faces expression limitations, such as over-smoothing and over-squeezing issues.
Some researchers have extended Transformers~\cite{2017Attention} to graph structures, combining attention mechanisms from natural language processing with graph neural network models. 
This approach avoids the expression limitations of GNNs and has shown promising results~\cite{2020A,2021Do,dwivedi2022benchmarking}.
However, real-world chemical molecules are exceedingly complex and diverse. For example, in the benzene ring, the large $\pi$-electron cloud is uniformly distributed over the six carbon atoms, exhibiting equivalent electron density. 
When a substituent is introduced onto the benzene ring, it can modulate the electron density on the ring either through inductive effects or conjugation. Consequently, the substituent, via both inductive and conjugative influences, can either increase or decrease the electron density on the benzene ring, inducing variations in the electron cloud distribution across the carbon atoms (see Figure ~\ref{Figure:00}(a)and(b)).

In response to these challenges, researchers have incorporated 3D information of molecular structures into Transformers, recognizing that drug properties and effects are heavily influenced by their molecular structures~\cite{2022ChemRxiv,2021Molformer}. However, capturing spatial information requires introducing additional units in the generator to account for Euclidean symmetries such as rotation, translation, and reflection, posing higher complexity and computational costs~\cite{ijms232113568}. 

Whether to effectively combine spatial information while preserving the simplicity and efficiency of graph network methods is an urgent issue that needs to be addressed in current research. Excitingly, mathematical invariants derived from differential geometry and algebraic topology are being viewed as descriptors for molecules. Curvature, as a basic concept in differential geometry, captures the intrinsic properties of a manifold surface~\cite{2015Ricci,2018Learning,2020Ricci}. 
Recently, advanced learning models based on these invariants have shown remarkable success in drug design, prediction of protein-ligand affinity, and forecasting the affinity of other organic and inorganic nanoscale particles, for their high level of abstraction and portability~\cite{Quantitative,2018Representability,2020Machine,wee2021ollivier,2022Unifying}. Interestingly, geometric and topological properties can effectively align with the intrinsic characteristics of quantum chemistry, 
 as illustrated in Figure~\ref{Figure:00}(c).
The 2-hop curvature assigns well-aligned weights to nodes, effectively aligning with the difference charge map of 1-Phenylpyrrole.

In this research, we introduce 
Curvature-Based Topology-Aware Graph Embedding (CTAGE) 
as a straightforward and effective descriptor for learning molecular representations. 
Our approach involves computing Discrete Ricci Curvature (DRC) across nodes at various cutoff radii, aimed at capturing intricate structural nuances within molecules, thereby enhancing the model's performance on intricate molecular graphs. 
This curvature-centric graph embedding method presents two pivotal advantages: (1) it offers a streamlined and efficient solution for versatile graph embedding, surpassing the constraints of conventional graph-theoretical encodings by adeptly recognizing structural diversity and alterations in molecular functional groups; (2) the adaptability in computing $k$-hop node curvature by adjusting the cutoff radius renders it suitable for various datasets across distinct task contexts.
The experimental results indicate that our method significantly enhances the model's performance in molecular prediction tasks on the Graph Transformer and GNNs models, without altering the underlying architecture.

\section{Related Work}
This section will present the research progress in molecular representation learning, spanning from one-dimensional to two-dimensional and three-dimensional spaces. It will also introduce some significant works closely related to our research direction.
\subsection{Molecular Deep Learning}
\paragraph{1D-representation learning} primarily involves directly utilizing SMILES strings to train natural language models, thereby extracting potential chemical property features of molecules~\cite{gomez2018automatic,2022ChemBERTa,xu2017seq2seq,honda2019smiles}. They typically rely on RNN sequence-to-sequence models and Transformer-based pre-trained models. They leverage abundant SMILES text data effectively. However, this approach could be limited due to the non-uniqueness and lack of interpretability in their results. 

\paragraph{2D-representation Learning} involves representing molecular data in graph structures, which can better express node and edge features. This method's topological structure also offers improved interpretability~\cite{2019Analyzing,velickovic2017graph,kearnes2016molecular,xiong2019pushing}. Some works also opt to use molecular images as auxiliary learning objects. Additionally, GNNs' message-passing mechanisms suffer from over-smoothing and over-squeezing, limiting their expressive power to some extent. The quality of graph embedding representations, particularly encoding node and edge features, significantly impacts the model's performance.

\paragraph{3D-representation Learning} Recent works in the field of drug discovery have reported promising results using graph Transformer models~\cite{2020Molecule,2021Do,2020GROVER,2020Graph}. These architectures facilitate the utilization of three-dimensional spatial information in molecules~\cite{2022ChemRxiv,liao2023equiformer}. However, this architecture constrains the model's ability to capture graph structural information as the self-attention computations are unaffected by node connectivity. Consequently, graph Transformer models often necessitate the design of additional absolute encoding or network modules to acquire graph structural and spatial information.

\subsection{Structural and Spatial Encoding}
The encoding methods for structural and spatial information within graphs represent a prominent research focus in molecular modeling tasks. Initially, approaches utilized graph theoretical information like Deepwalk PE~\cite{perozzi2014deepwalk}, Random walk PE~\cite{li2020deepergcn}, Shortest Path Distances PE~\cite{li2020distance}, and  Centrality Encoding~\cite{2021Do}. However, these methods couldn't further distinguish structural differences within graphs, often necessitating additional extraction of local structural information from nodes.

Another approach involves considering higher-order three-dimensional information by employing E(3)~\cite{satorras2021n} or SE(3)~\cite{thomas2018tensor} networks to integrate spatial information into models. These methods require designing additional network modules and involve more complex training processes. 
Therefore, this paper explores a more concise and efficient Embedding method for capturing both structural and spatial information within graphs: Curvature-Based Topology-Aware Graph Embedding (CTAGE).

\subsection{Discrete Ricci Curvature}
The discrete curvature is taken as a measure of the graph structure on the manifold, and this measure does not change its topology~\cite{2013Stochastic}. It describes the inter-correlation case of the neighborhoods between a pair of nodes.
Most of the previous work on graph curvature in combining graph neural networks was done to optimize the data structure. Specifically, different curvature calculation methods are used to smooth the curvature of the graph data. In addition, the curvature on the graph is modified by adding the links or modifying the weights of the nodes on the graph, so that the curvature of the overall data tends to smooth, to enhance the network performance or alleviate the over-squashing~\cite{2020ICLR,2021Understanding}.
Current mainstream approaches are Ollivier Ricci curvature~\cite{ollivier2009ricci,Lin2010Ricci}and Forman curvature ~\cite{Forman2003}.

Among them, Ollivier Ricci Curvature has been proven to be very successful in the communication network, but its calculation process involves the optimal transmission calculation problem between nodes. 
It has high complexity and is not suitable for graph prediction and graph regression problems with huge data volumes. 
However, the calculation of Forman Ricci curvature is relatively simple and applicable to both directed and undirected weighted graphs. 
It is well-suited for studying interaction relationship networks, protein structure networks, and molecular networks ~\cite{2016Forman}. 

Hence, in our pursuit of a thorough evaluation of curvature attributes and computational intricacies, we opt for employing Forman Ricci curvature to more effectively encapsulate structural information within tasks related to predicting molecular properties.

\section{Preliminaries}
Before introducing the proposed node Curvature Embedding,
we recall some technical details of the computational methods for the Forman Ricci Curvature~\cite{Forman2003}. 

\paragraph{Forman-Ricci Curvature} 
While Ollivier Ricci curvature stands out in the forefront, it is considered to be more suitable for studying information transfer in communication networks and might have possible limitations in interaction networks, such as inter-protein interaction networks and molecular networks~\cite{2016Forman}. 
Therefore, we introduce the Forman curvature as an alternative metric. The calculation is based on the edges and is specifically applicable to undirected and weighted networks. The computation is defined as follows:
\begin{equation}
\begin{aligned}
\mathcal{F}(e)& = 
w_e \left(
        \frac{w_{v_i}}{w_e} + \frac{w_{v_j}}{w_e}\right)  \\
&-w_e\left(\sum_{e_{v_i} \sim e, \ e_{v_j} \sim e} \left[
            \frac{w_{v_i}}{\sqrt{w_e w_{e_{v_i}}}} + \frac{w_{v_j}}{\sqrt{w_e w_{e_{v_j}}}}
        \right]
    \right)
\end{aligned}
\label{Edge_forman}
\end{equation}
where $e$ is the edge to be calculated, connected to $v_i$ and $v_j$, 
$w_e$ is the weight of edge $e$, 
$w_{v_i}$ and $w_{v_j}$ are the weights of two nodes, 
$e_{v_i}\sim e$, $e_{v_j}\sim e$ are sets of edges connected to $e$ via nodes $v_i$ and $v_j$. 

Although the computation of Forman-Ricci curvature is defined based on edges, it can be naturally extended to nodes.

\paragraph{Discrete Ricci Node Curvature}
The Discrete Ricci Curvature of a node can be computed by averaging all curvatures of the edges $e\sim v$ that are connected to the node $v$~\cite{2016Forman}, as shown below:
\begin{equation} 
    \mathcal{F}(v)=\frac{1}{\deg{(v)}}\sum _{e\sim v}\mathcal{F}(e)
    \label{node_curvature}
\end{equation}
where $\deg(v)$ is the degree of node $v$.

\section{Methodology}
In this section, we introduce the method for calculating discrete curvature on molecular graphs, the transformation of negative node curvature values, and how to incorporate node curvature embedding into graph network models. We then propose the $K$-hop node discrete Ricci curvature.
\subsection{Calculating Curvature on Molecular Graph}
Given a specific molecular data (SMILES string) with $n$ atoms, we can convert it into an undirected graph $G= (V, E)$, where $V = \{v_1, v_2, \cdots, v_n\}$ denote the atoms of the molecular and $E$ denotes the bond between two atoms. 
Then the node features of $V$ can be described as $X \in \mathbb{R}^{n \times 9}, (X = \{x_1, x_2, \cdots, x_n\})$ and the edge features are $H= {h_1, h_2, h_3}$~\cite{landrum2013rdkit}. Since the molecular graphs generated by this method are all undirected and unweighted, it is necessary to redefine ~\cref{Edge_forman}.

Let all nodes in the graph have the same mass and all edge weights are equal to 1. Then, the edge curvature value calculation~\cref{Edge_forman} for the edge $e$ between nodes $v_1$ and $v_2$ is redefined:
\begin{equation}
    \mathcal{F}(e)=4-\sum _{v\sim e}{\deg{(v)}}
    \label{easy_forman}
\end{equation}
where $v\sim e$ is the set of nodes connected to $e$. 

Similarly, when the generated molecular graphs are 3D, the weight of the edges changes according to the distances between the connected atoms. Then, let $v_i$ and $v_j$ be two nodes with three-dimensional coordinates $(x_i, y_i, z_i)$ and $(x_j, y_j, z_j)$, respectively. The edge weight $w_e$ can then be defined as the Euclidean distance between these two nodes:
\begin{equation}
    w_e = \sqrt{(x_j - x_i)^2 + (y_j - y_i)^2 + (z_j - z_i)^2}
\end{equation}

In this case, the ~\cref{Edge_forman} can be expressed in the following form:
\begin{equation}
\mathcal{F}(e) = 2 -\left(\sum_{e_{v_i} \sim e, \ e_{v_j} \sim e} \left[
            \sqrt{\frac{w_e}{ w_{e_{v_i}}}} + \sqrt{\frac{w_e}{ w_{e_{v_j}}}} \right] \right)
            \label{3D_forman}
\end{equation}

When dealing with tasks such as molecular oscillation, it is often necessary to enhance the model's recognition accuracy through subtle changes in chemical bonds. Then \cref{3D_forman} is more suitable for addressing such issues. In this work, we adopted the approach presented in \cref{easy_forman} because the distance interval between atoms is approximately [1.2Å, 1.6Å], allowing different edge weights to be considered equivalent.

\paragraph{Negative Curvature Transformation} Due to the inevitable presence of negative curvature values in the computation, incorporating negative curvature excessively as an additional signal in the network might potentially affect the performance of \emph{Curvature Encoding} and feature embedding. To address this, we designed multiple mapping functions to project the curvature onto a non-negative space. The mapping function that yields the best performance is as follows:
\begin{equation}
    \mathcal{F}(v) = \left( \frac{cur(v)-cur_{\min}}{cur_{\max}-cur_{\min}} \right )  \label{MM_map}
\end{equation}
and the experimental results of all functions can be found in the supplementary materials. 
\begin{figure}[htbp]
 \centering
 \includegraphics[width=8.5cm]{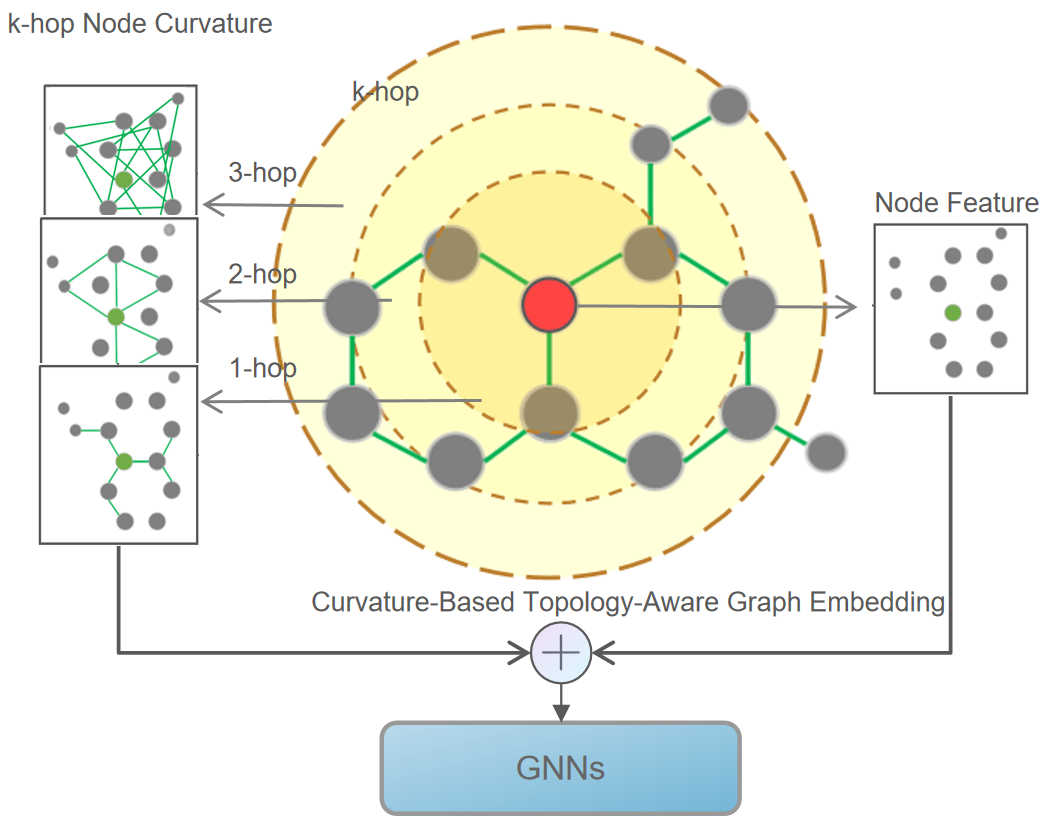}
 \caption{Encodes $k-$hop node curvature information of atoms in molecules at different cutoff radii into node features.}
\label{Figure:01}
\end{figure} 
\subsection{Embedding Curvature Information to Model}
The current mainstream architectures in graph neural networks consist of two types: one relies on message-passing mechanisms in GNNs, while the other utilizes self-attention mechanisms in Graph Transformers. To validate the universality of our embedding method, we conducted experiments with two classic models, Graph Convolutional Networks (GCN)~\cite{kipf2016semi} and Graphormer~\cite{2021Do}, selected from each of these two architectures.

\paragraph{CTAGE on Graphormer}
The degree centrality introduced in Graphormer~\cite{2021Do} and derived from social networks may not be as applicable in the field of chemistry. 
In chemistry, the significance of atoms is often determined by their bonding flexibility and their capacity to connect with other functional groups. 
The degree of carbon atoms in a molecule is always equal to '3' because carbon atoms typically form four covalent bonds. However, the functional roles and influences of carbon atoms connecting different functional groups in the molecule are significantly different.
Therefore, introducing CTAGE into the Graph Molecular model becomes necessary to generate more scientifically accurate results in the field of biochemical molecules.
We integrate the atomic node curvatures obtained from subgraphs with different hop distances, where the curvatures calculated from subgraphs with larger hop distances have less impact. Subsequently, we convert the aggregated $k$-hop node curvatures into positional encoding form through an embedding operation, as illustrated below:
\begin{equation}
        z_{cur}(v^k_i)=\text{Embedding} \left[ \text{dict} \left \{ \mathcal{F}(v^k_i) \right \}   \right]
\end{equation}
where $z_{cur(v_i)} \in \mathbb{R}^n$ is a learnable embedding vector and $\mathcal{F}(v^k_i)$ is the curvature of the node $v_i$ in the $k$-hop graph. The parameter "$k$" represents the number of hops for the $k$-hop subgraph, and "\text{dict}" represents a dictionary mapping, which means mapping the curvature information to the input format required by the network or the required position encoding format. The mapping range is usually determined by the maximum number of nodes or the dimensionality of node features in the graph dataset.

Then, the form of curvature-driven centrality encoding is as follows:
\begin{equation}
    h_i = x_i + z_{cur(v_i)}.
    \label{cent}
\end{equation}

\paragraph{CTAGE on GCN}
GCN is adept at extracting node features in a graph. A two-layer deep GCN exhibits remarkably high performance~\cite{kipf2016semi}. We selected the GCN backbone from the work of~\cite{li2021dgl} as our research subject, where node feature encodings and connectivity are from the model's input. Continuing the embedding research in Graphormer, we modified the encoding generation block for Atom degree excluding hydrogen atoms in its node descriptors.
Specifically, firstly applying~\cref{MM_map} to map and convert the node curvature values (excluding hydrogen atoms) into one-hot encoded form.
Then, utilizing this encoding to replace the original encoding in the node feature vector $x \in \mathbb{R}^{74}$, replacing dimensions from the 43rd to the 53rd.

\subsection{$k$-hop Node Curvature} 
Although curvature information has a broader scope than degree information, it still only reflects local structural information within a relatively small range of nodes. To overcome this limitation, we introduce the concept of $k$-hop curvature, involving the establishment of additional relational networks based on various cutoff radii from the central atom (e.g., the red point in Figure~\ref{Figure:01}). This process generates $k$-hop subgraphs, allowing the utilization of node curvature within these subgraphs to quantify extensive node structural information. The algorithmic workflow is outlined as follows.
\begin{algorithm}[H]
    \caption{Computation of $K$-hop Ricci Curvature}
    \label{alg:AOA}
    \renewcommand{\algorithmicrequire}{\textbf{Input:}}
    \renewcommand{\algorithmicensure}{\textbf{Output:}}
    \begin{algorithmic}[1]
        \REQUIRE A graph $G <V,E>$, number of nodes $n$, temp degree vector $(d_1,d_2,\cdots,d_n), d_i = 0$, SPD $P$.  
        \ENSURE $k$-hop node Curvature $(\mathcal{F}(v^k_0),\mathcal{F}(v^k_1),\cdots,\mathcal{F}(v^k_n))$  
        \STATE Rebuild the $k$-hop Graph of molecules.
        \FOR{$i=1$ to $n$}
            \FOR{$j=i$ to $n$}
                \IF{$P_{ij} = k$}
                \STATE $d_i++$ 
                \STATE $d_j++$
                \STATE add $e_{ij}$ to $E$
                \ENDIF
            \ENDFOR
        \ENDFOR
        \STATE Initialization of all weights on nodes and edges to 1.
        \FOR{each $v_i \in V$}
            \STATE Compute the $k$-hop DRC $\mathcal{F}(v^k_i)$ of the node $v^k_i$.\\ $\mathcal{F}(v^k_i) = \frac{1}{d_i}\sum_{e_v\sim v}\mathcal{F}(e_v)$;
        \ENDFOR
    \end{algorithmic}
\end{algorithm}

\section{Experimental Results}
In the upcoming experiments, we will be conducting molecular property prediction tests using node Curvature Embedding.
The codes are available in Supplementary Material.

\subsection{Datasets and Baselines}
\paragraph{Datasets} 
In this paper, we conducted experiments using MoleculeNet~\cite{2018MoleculeNet}, which serves as a benchmark designed for evaluating machine learning methods in predicting molecular properties. The dataset underwent scaffold splitting during preprocessing~\cite{bemis1996properties}. Root Mean Square Error (RMSE) was employed for molecular regression tasks, and Receiver Operating Characteristic Area Under the Curve (ROC-AUC) was used for molecular prediction tasks as evaluation metrics. Test performance was determined based on the model that exhibited the best results in the validation setting.

Our experiments encompassed various tasks and datasets, including ESOL, FreeSolv, Lipophilicity~\cite{2018MoleculeNet}, Blood-brain barrier permeability (BBBP), BACE~\cite{Denny2016Computational}, and ClinTox~\cite{Kaitlyn2016A}. For detailed information about these datasets, please refer to the Supplementary Material.

\paragraph{Baselines} 
On the MoleculeNet~\cite{2018MoleculeNet} dataset, we independently assessed the effectiveness of CTAGE using two models: GCN and Graphormer, and conducted separate comparisons based on their respective architectural types. For the GCN comparison, we employed GNNs based on the Message Passing, while for the Graphormer, we utilized GNNs based on the Transformer architecture. Among them, Uni-Mol~\cite{2022ChemRxiv}, GROVER~\cite{2020GROVER}, and AttentiveFP~\cite{xiong2019pushing} are pretraining frameworks. In particular,  Uni-Mol also considers the 3D information.

\begin{table*}[h]
\begin{center}
{\fontsize{8pt}{11pt}\selectfont 
\begin{tabular}{llll|lll}
\hline
            & BACE  & BBBP  & ClinTox                    & EOSL  & FreeSolv & Lipo  \\
            & ROC-AUC   & ROC-AUC   & ROC-AUC   & RMSE  & RMSE     & RMSE  \\ \hline
D-MPNN~\cite{2019Analyzing}      & 0.857(0.057) & \textbf{0.697(0.058)} & 0.906(0.036) & 1.050(0.008) & 2.082(0.082)    &  \textbf{0.683(0.016)} \\
GAT~\cite{velickovic2017graph}      &0.697(0.064) &0.662(0.026)  &0.585(0.036)  &1.556(0.085)  &3.559(0.050)  &1.021(0.029)  \\
AttentiveFP~\cite{xiong2019pushing}      &0.784(0.022)  &0.643(0.018)  &0.847(0.003)  &\textbf{0.877(0.029)}  &\textbf{2.073(0.183)} &0.721(0.001)  \\
GCN~\cite{kipf2016semi}         & 0.832(0.021) & 0.692(0.042) & 0.904(0.028) & 1.118(0.314) & 2.521(0.348)    & 0.792(0.035) \\
GCN$^{1hop}_{CTAGE}$        & \underline{\textbf{0.865(0.015)}} & 0.689(0.022) & 0.894(0.028) & 0.917(0.076) & 2.279(0.242)    & \underline{0.767(0.02)} \\
GCN$^{2hop}_{CTAGE}$       & 0.803(0.031) & 0.676(0.011) & \underline{\textbf{0.912(0.047)}} & \underline{0.890(0.029)} & \underline{2.192(0.184)}    & 0.812(0.025) \\
GCN$^{3hop}_{CTAGE}$        & 0.801(0.023) & \underline{\textbf{0.697(0.051)}} & 0.905(0.023) & 1.007(0.401) & 2.572(0.199)    & 0.787(0.033) \\ \hline
GROVER$_{base}$~\cite{2020GROVER}  & 0.826(0.007) & 0.700(0.001) & 0.812(0.030) & 0.983(0.090) & 2.176(0.052)  & 0.817(0.008) \\
GROVER$_{large}$~\cite{2020GROVER}  & 0.810(0.010) & 0.695(0.001) & 0.762(0.037) & 0.895(0.017) & 2.272(0.051)    & 0.823(0.046) \\
Uni-Mol~\cite{2022ChemRxiv}      &\textbf{0.857(0.002)}  &0,729(0.006)  &0.919(0.018)  &\textbf{0.788(0.029)}  &\textbf{1.620(0.035)}  &\textbf{0.603(0.010)}  \\
Graphormer~\cite{2021Do}  & 0.811(0.047) & 0.712(0.029) & 0.943(0.035) & 0.860(0.027) & 2.214(0.175)    & 0.725(0.046) \\
Graphormer$^{1hop}_{CTAGE}$ & 0.821(0.053) & 0.727(0.027) & 0.944(0.032) & 0.841(0.036) & \underline{2.208(0.29)}    & 0.682(0.043) \\
Graphormer$^{2hop}_{CTAGE}$ & \underline{0.838(0.057)} & \underline{\textbf{0.730(0.043)}} & \underline{\textbf{0.954(0.027)}} & \underline{0.837(0.038)} & 2.289(0.235)    & \underline{0.675(0.028)} \\
Graphormer$^{3hop}_{CTAGE}$ & 0,826(0.046) & 0.711(0.055) & 0.946(0.055) & 0.854(0.041) & 2.324(0.311)    & 0.733(0.057) \\ \hline
\end{tabular}}
\end{center}
\caption{The results comparison. The optimal results are shown in \textbf{bold}, and the optimal results of ours are shown in \underline{\textit{underline}}.}
\label{tab:01}
\end{table*}

\subsection{Implementation Details}
In this section, we provide details on the hyperparameters and training settings employed in our experiment. When utilizing Graphormer as the backbone network, we configured the network with a depth of 10, 32 self-attention heads, and 128 channels in the hidden layer. Given the relatively small size of the molecular graph, we set the embedding dropout ratio to 0.0, and the learning rate was fixed at 3e-4. These parameter values were carefully selected based on optimal outcomes derived from experimental exploration.

For the GCN model, the network comprises 2 layers, a dropout ratio of 0.05, 256 hidden units, and a learning rate of 2e-2, as per the specifications provided by ~\cite{li2021dgl}. Adam was uniformly used for all models, and each experiment was conducted over five iterations with different random seeds. Additional experimental details and results are available in the Appendix.
%
\subsection{Results for Molecular Property Prediction}
The experimental results of two CTAGE models and competitive baselines are presented in Table~\ref{tab:01}. This indicates that, without relying on 3D information and without introducing additional training units, CTAGE can significantly enhance the performance of graph network models in predicting molecular properties. This is because the k-hop node curvature information better describes the molecular topological structure, enabling the model to capture the chemical information inherent in molecular structures. Although we did not outperform the SOTA on all datasets, considering that Uni-Mol and AttentiveFP were pre-trained on a large amount of data and Uni-Mol additionally utilized 3D information of molecules, it is reasonable that our results may not surpass all models.

In Graphormer, 2-hop CTAGE shows a noticeable enhancement, indicating that 2-hop node curvature provides more detailed topological information without losing generality. On the other hand, the computation of 3-hop node curvature takes into account a broader range of structural information, a practice that may lead to overfitting. This is observed in our experiments where we witnessed a significant increase in accuracy on the training set, while the accuracy on the testing set tended to decrease.

\begin{figure}[htbp]
 \centering
 \includegraphics[width=8.5cm]{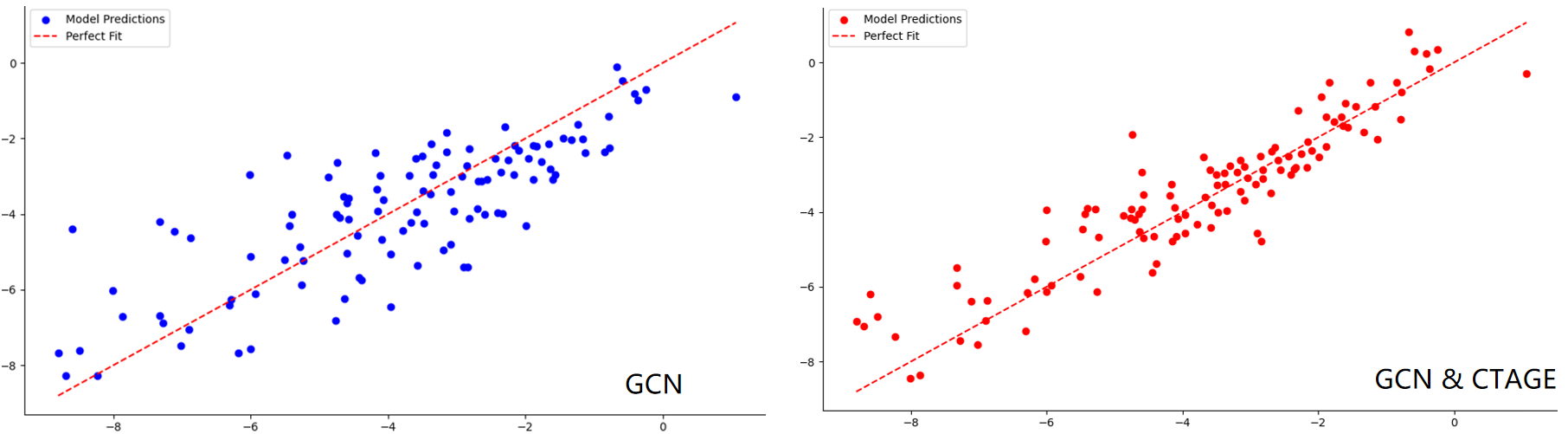}
 \caption{Comparison of regression scatter plots between GCN and GCN \& CTAGE (2-hop node curvature) on the ESOL test set.}
\label{Figure:03}
\end{figure}

\begin{figure}[htbp]
 \centering
 \includegraphics[width=8.5cm]{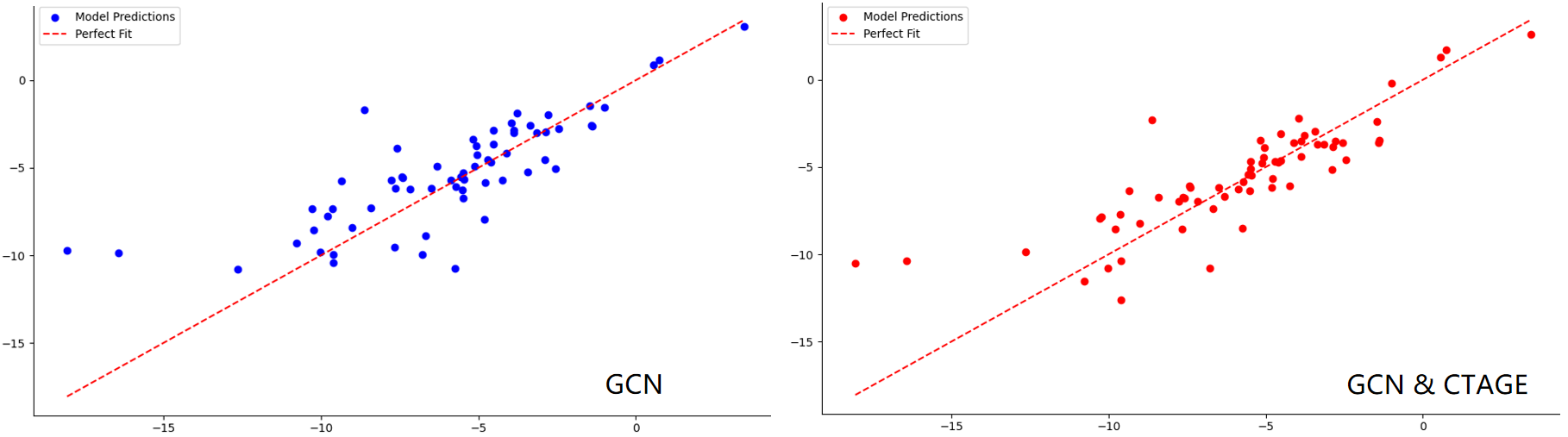}
 \caption{Comparison of regression scatter plots between GCN and GCN \& CTAGE (2-hop node curvature) on the FreeSolv test set.}
\label{Figure:04}
\end{figure}

\begin{figure}[htbp]
 \centering
 \includegraphics[width=7cm]{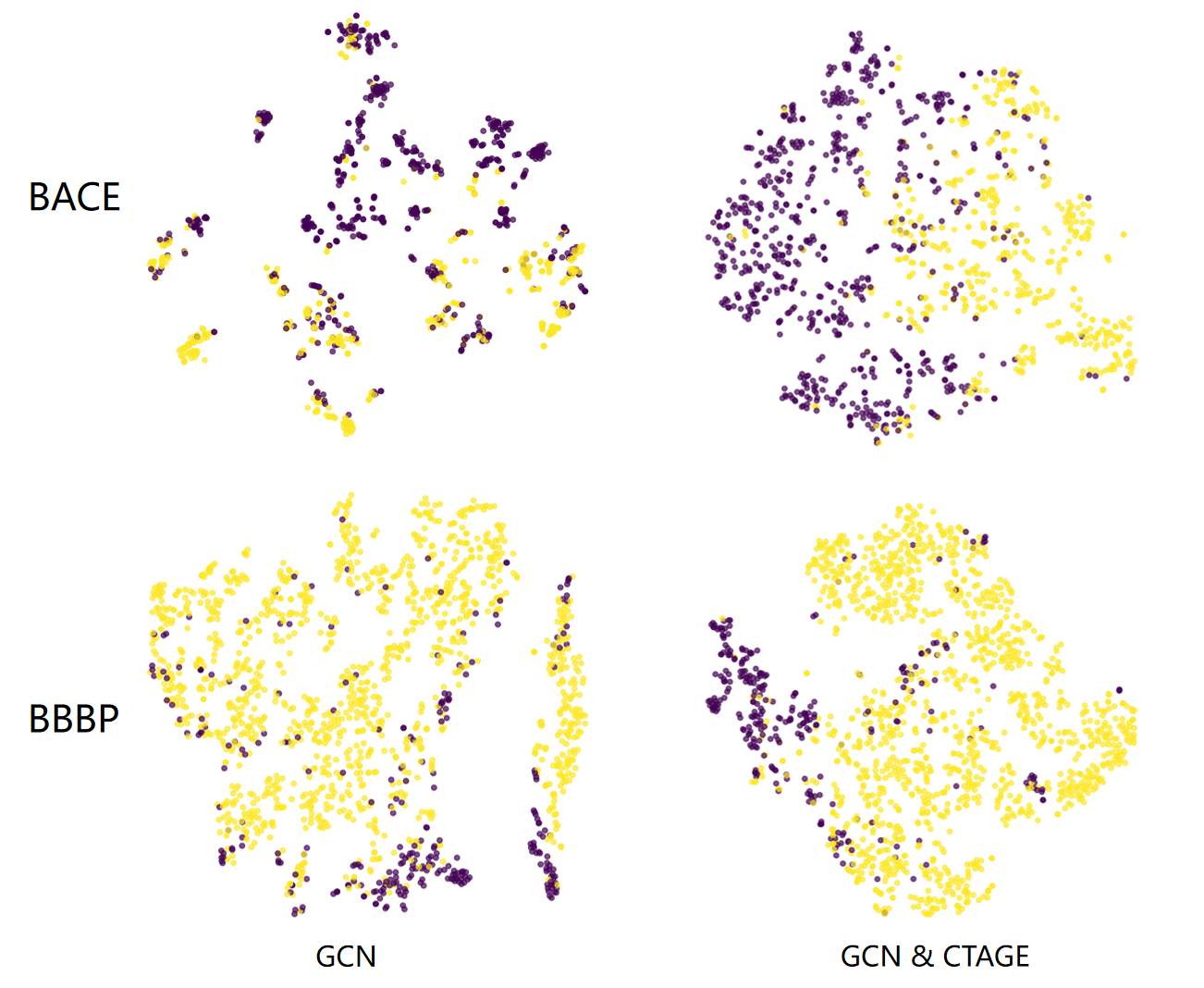}
 \caption{Visualization of the latent space of GCN on BACE and BBBP datasets. Using t-SNE to map extracted molecular features into 2D coordinate points, assigning different colors to points based on the positive or negative values of the labels.}
\label{Figure:05}
\end{figure}

In the GCN network, due to the limited dimensionality of the mapping in GCN~\cite{li2021dgl}, which is only 11 dimensions, some curvature information gets compressed. This results in less-than-ideal experimental outcomes in certain cases. However, overall, it is still evident that $k$-hop node curvature information provides a more effective improvement to the model compared to the original degree information. The feature distributions of the BACE and BBBP datasets were visualized using t-SNE~\cite{van2008visualizing}, as depicted in Figure~\ref{Figure:05}. It can be observed that CTAGE allows the model to better learn the relationships between structure and molecular chemical properties. Different categories of data occupy corresponding regions in the latent space. We also visualized the prediction results on the ESOL and FreeSolv test sets through scatter plots in Figure~\ref{Figure:03} and Figure~\ref{Figure:04}, indicating that node curvature information is more effective in conveying topological information within molecular structures. Additionally, these pieces of information are beneficial for assessing their chemical properties.


\begin{figure*}[htbp]
 \centering
 \includegraphics[width=15cm]{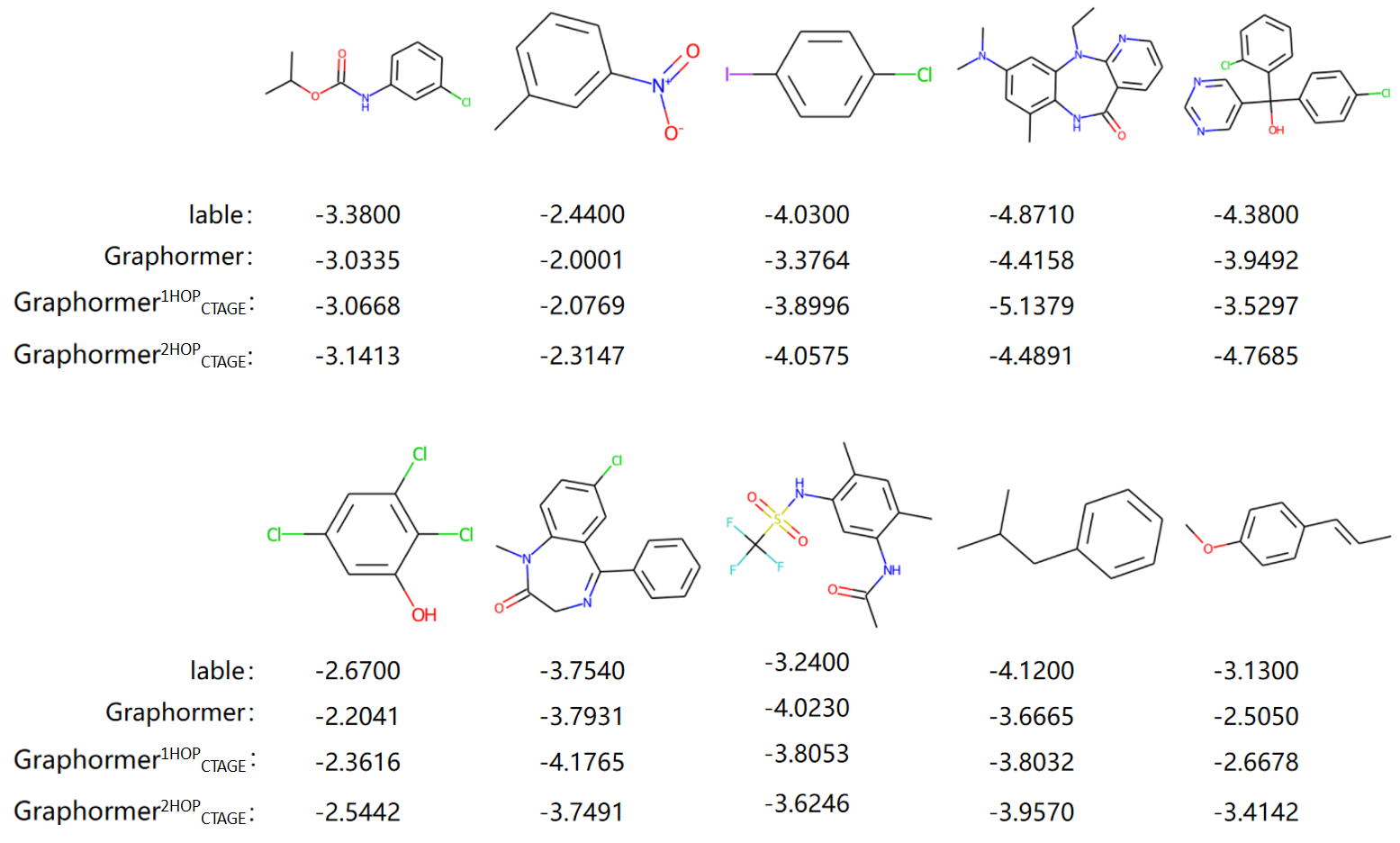}
 \caption{A statistical analysis of benzene derivatives from the ESOL dataset. Comparison of prediction results between Graphormer and CTAGE with 1-hop and 2-hop node Curvature applied separately.}
\label{Figure:02}
\end{figure*}

\begin{table}[htbp]
    \centering
    \caption{Comparison between Centrality, $1$-hop Node-Curvature and $2$-hop Node-Curvature Embedding in Graphormer}
    {\fontsize{7.5pt}{10pt}\selectfont 
\begin{tabular}{@{}lllc@{}}
\toprule
Dataset                                    & Molecule Size                                                & Method                                                                                 & AUC ($\uparrow$)                \\ \midrule
\multicolumn{1}{l|}{\multirow{6}{*}{BACE}} & \multicolumn{1}{l|}{\multirow{3}{*}{atom\_num\textless{}34}} & \multicolumn{1}{l|}{Graphormer}                          & 0.798                           \\
\multicolumn{1}{l|}{}                      & \multicolumn{1}{l|}{}                                        & \multicolumn{1}{l|}{Graphormer$^{1hop}_{CTAGE}$} & 0.854 \\ \multicolumn{1}{l|}{}                      & \multicolumn{1}{l|}{}                                        & \multicolumn{1}{l|}{Graphormer$^{2hop}_{CTAGE}$} & \textbf{0.915} \\ \cmidrule(l){2-4} 
\multicolumn{1}{l|}{}                      & \multicolumn{1}{l|}{\multirow{3}{*}{atom\_num$\geq$34}}      & \multicolumn{1}{l|}{Graphormer}                          & 0.674                         \\
\multicolumn{1}{l|}{}                      & \multicolumn{1}{l|}{}                                        & \multicolumn{1}{l|}{Graphormer$^{1hop}_{CTAGE}$} & 0.788                           \\  \multicolumn{1}{l|}{}                      & \multicolumn{1}{l|}{}                                        & \multicolumn{1}{l|}{Graphormer$^{2hop}_{CTAGE}$} & \textbf{0.852} \\ \midrule
\multicolumn{1}{l|}{\multirow{6}{*}{BBBP}} & \multicolumn{1}{l|}{\multirow{3}{*}{atom\_num\textless{}18}} & \multicolumn{1}{l|}{Graphormer}                          & 0.818                         \\
\multicolumn{1}{l|}{}                      & \multicolumn{1}{l|}{}                                        & \multicolumn{1}{l|}{Graphormer$^{1hop}_{CTAGE}$} & 0.835 \\  \multicolumn{1}{l|}{}                      & \multicolumn{1}{l|}{}                                        & \multicolumn{1}{l|}{Graphormer$^{2hop}_{CTAGE}$} & \textbf{0.864} \\ \cmidrule(l){2-4} 
\multicolumn{1}{l|}{}                      & \multicolumn{1}{l|}{\multirow{3}{*}{atom\_num$\geq$18}}      & \multicolumn{1}{l|}{Graphormer}                          & 0.785                          \\
\multicolumn{1}{l|}{}                      & \multicolumn{1}{l|}{}                                        & \multicolumn{1}{l|}{Graphormer$^{1hop}_{CTAGE}$} & 0.788                           \\ \multicolumn{1}{l|}{}                      & \multicolumn{1}{l|}{}                                        & \multicolumn{1}{l|}{Graphormer$^{2hop}_{CTAGE}$} & \textbf{0.824} \\ \bottomrule
\end{tabular}}
\label{tab:04}
\end{table}

\subsection{Analysis}
To explore the potential optimization of curvature information (Forman curvature) in molecular tasks of varying scales, we utilized the average number of atoms in the dataset as a criterion to divide the test sets of the BACE and BBBP datasets. 
The experimental results are presented in Table~\ref{tab:04}. 
Through analyzing these results, we observed that in the BACE dataset, as the graph size increased, Graphormer exhibited a gradual decline in competitiveness. 
We hypothesize that as the molecular scale increases, the structure also becomes more complex, which adds difficulty to the prediction task. The model struggles to capture the interdependence between nodes in complex structures fully.
Therefore, whether it is centrality encoding or CTAGE, their limited receptive fields are insufficient to capture the information gain brought by complex structures fully.

To address this issue, we partitioned the test set into different subsets based on the number of atoms in the molecular graphs. Testing was conducted on these subsets with varying scales of atom numbers. For subsets with more nodes and more complex structures, we performed a $2$-hop subgraph reconstruction and computed node curvature on these reconstructed graphs, referred to as $2$-hop curvature. Experimental results demonstrated that introducing node curvature information from the $2$-hop subgraphs further improved the model's performance, as shown in Table~\ref{tab:04}. We also collected predictions on the ESOL test set and found that curvature contributes more to the model's expression on molecules with benzene rings, see Figure~\ref{Figure:02}. Additionally, we observed a certain correlation between the node curvature in these $2$-hop subgraphs and the original node curvature, with detailed information explained in the supplementary materials. information explained in the supplementary materials.

\section{Discussion and Future Work}
Our work, based on the current mainstream architectures of 
graph convolutional networks and graph transformers,
evaluates the performance of Curvature-Based Topology-Aware Graph Embedding (CTAGE) across molecular datasets and achieves robust and competitive results. 
This success is primarily attributed to the introduction of curvature, providing additional topological information that highlights the correlation between molecular structure and properties. It enriches the structural information in the encoding of node features, allocating attention weights more efficiently and accurately.

However, the current CTAGE formulation does not explicitly consider the weight impact of atoms and chemical bonds, as well as the spatial distances of three-dimensional nodes. Consequently, it cannot distinguish structural differences in chiral molecules and fails to effectively express the structural information of crystal materials with a uniform lattice system. To address this limitation, our plans involve incorporating more chemical information along with three-dimensional spatial information into the encoding calculations. This is expected to enhance CTAGE's ability to describe the topological information of chemical graph data more effectively.

While extending the perceptual range of curvature information through $k$-hop curvature, selecting an appropriate truncation radius remains a challenge in datasets with diverse task attributes. Therefore, we will continue to explore the deeper correlations between curvature information and molecules. Additionally, we plan to integrate curvature information with a one-dimensional drug discovery model, constructing a SMILES vocabulary that aligns more closely with chemical intuition based on molecular structural information. 
This integration aims to advance the application of LLM in the field of drug discovery.


\newpage
\bibliographystyle{named}

\bibliography{ijcai24}

\end{document}